\documentclass[sigconf,anonymous=false]{acmart}
\def\BibTeX{{\rm B\kern-.05em{\sc i\kern-.025em b}\kern-.08emT\kern-.1667em\lower.7ex\hbox{E}\kern-.125emX}}
    
\renewcommand\footnotetextcopyrightpermission[1]{}
\usepackage{nicefrac}
\usepackage{siunitx}
\usepackage{array,framed}
\usepackage{booktabs}
\usepackage{float}
\usepackage{
  color,
  float,
  epsfig,
  wrapfig,
  graphics,
  graphicx,
  subcaption
}
\usepackage{textcomp}
\usepackage{setspace}
\usepackage{latexsym,fancyhdr,url}
\usepackage{enumerate}
\usepackage{algorithm2e}
\usepackage{algpseudocode}
\usepackage{graphics}
\usepackage{xparse} 
\usepackage{xspace}
\usepackage{multirow}
\usepackage{csvsimple}
\usepackage{balance}
\usepackage{caption}
 \hypersetup{
     colorlinks=true,
     linkcolor=blue,
     filecolor=magenta,      
     urlcolor=cyan,
     pdftitle={Overleaf Example},
     }

\usepackage{
  tikz,
  pgfplots,
  pgfplotstable
}

\pgfplotsset{compat=1.9}

\usepackage{mathtools,}

\DeclareMathAlphabet{\mathcal}{OMS}{cmsy}{m}{n}

\begin{document}

\def\thetitle{Sisyphus: A Cautionary Tale of Using Low-Degree Polynomial Activations in Privacy-Preserving Deep Learning}
\title{\thetitle}

\author{Karthik Garimella, Nandan Kumar Jha, Brandon Reagen \\
New York University \\
\texttt{\{kg2383,nj2049,bjr5\}@nyu.edu}}

\date{}

\begin{abstract}
Privacy concerns in client-server machine learning have given rise to private inference (PI), where neural inference occurs directly on encrypted inputs. PI protects clients' personal data and the server's intellectual property. A common practice in PI is to use garbled circuits to compute nonlinear functions privately, namely ReLUs. However, garbled circuits suffer from high storage, bandwidth, and latency costs. 
To mitigate these issues, PI-friendly polynomial activation functions have been employed to replace ReLU. 
In this work, we ask: Is it feasible to substitute all ReLUs with low-degree polynomial activation functions for building deep, privacy-friendly neural networks? 
We explore this question by analyzing the challenges of substituting ReLUs with polynomials, starting with simple drop-and-replace solutions to novel, more involved replace-and-retrain strategies. 
We examine the limitations of each method and provide commentary on the use of polynomial activation functions for PI. We find all evaluated solutions suffer from the \textit{escaping activation} problem: forward activation values inevitably begin to expand at an exponential rate away from stable regions of the polynomials, which leads to exploding values (NaNs) or poor approximations.

\end{abstract}

\maketitle

\section{Introduction}
\label{sec:intro}

The growing adoption of Machine Learning as a Service (MLaaS) \cite{hunt2018chiron} has given rise to privacy concerns of clients' personal data and the intellectual property (i.e., trained models) of service providers. To address these concerns, techniques such as different privacy \cite{dwork2006differential,abadi2016deep}, federated learning \cite{konevcny2016federated,bonawitz2019towards}, secure enclaves \cite{costan2016intel,tramer2018slalom}, homomorphic encryption (HE) \cite{gentry2009fully}, and multiparty computation (MPC) \cite{shamir1979share} aim to prevent both the server from accessing the client's sensitive data and the client from learning the server's model. 
One area of study within privacy-preserving machine learning (PPML) attempts to perform inference directly on encrypted data 
using either HE \cite{gilad2016cryptonets,mohassel2017secureml,sanyal2018tapas} or MPC-based techniques such as Secret Sharing (SS) \cite{mohassel2018aby3,riazi2018chameleon,riazi2019xonn,rouhani2018deepsecure,rachuri2019trident,patra2020blaze,chaudhari2019astra,chandran2019ezpc}. Common PI protocols employ HE/SS for processing linear operations (e.g., convolutions and fully connected layers) and garbled circuits for nonlinear operations (e.g., ReLU and maxpool) \cite{liu2017oblivious,juvekar2018gazelle,mishra2020delphi,rathee2020cryptflow2,SAFENET,jha2021deepreduce}.


Garbled circuits are a major source of inefficiency when performing PI for the following reasons: \textbf{(1)} in PI, unlike plaintext inference, ReLU garbled circuits dominate the runtime and can be orders of magnitude more costly than linear layers computed with SS \cite{cryptonas,mishra2020delphi};
\textbf{(2)} a single ReLU operation using garbled circuits requires 17.5 KB of data storage and communication, and a single inference on state-of-the-art DNNs (such as ResNet50 \cite{he2016deep}) requires  millions of ReLU computations that leads to hundreds of GiB of data storage and communication \cite{rathee2020cryptflow2}.
These inefficiencies exist for variants of ReLU such as leaky ReLU \cite{maas2013rectifier}, parametric ReLU \cite{he2015delving}, RReLU \cite{xu2015empirical}, CReLU \cite{shang2016understanding}, and the recently proposed DY-ReLU \cite{chen2020dynamic}. Furthermore, storage and latency costs of GCs are exacerbated when used to compute more expressive and complex activation functions such as ELU \cite{clevert2015fast}, SELU \cite{klambauer2017self}, Swish \cite{ramachandran2017searching}, GELU \cite{hendrycks2016gaussian} and Mish \cite{misra2019mish}.

\begin{table} [t] \vspace{1cm}
\caption{Effectiveness of polynomial activations proposed for ciphertext training/inference on MNIST, CIFAR-10 (C10), CIFAR-100 (C100), and TinyImageNet (Tiny) datasets. * denotes the full ImageNet dataset.}
\label{tab:MethodComph}
\centering 
\resizebox{0.49\textwidth}{!}{
\begin{tabular}{ccccccc} \toprule
\multirow{2}{*}{Method} & \multirow{2}{*}{Degree} & \multirow{2}{*}{Partial/Full} & \multicolumn{4}{c}{Datasets} \\ 
\cmidrule(lr{0.5em}){4-7}  
& & &  MNIST & C10 & C100 & Tiny \\ \toprule
CryptoNet \cite{gilad2016cryptonets} & 2 & Full & Y & N & N & N  \\ 
Lookup Table \cite{thaine2019efficient} & 2& Full & Y & N & N & N  \\ 
Polyfit \cite{chabanne2017privacy} & 2, 4, 6  & Full & Y & N & N & N  \\ 
SecureML \cite{mohassel2017secureml} & 2  & Full & Y & N & N & N  \\ 
FCryptoNet \cite{fastercryptonets} & 2 & Full & Y & Y & N & N  \\ 
CryptoDL \cite{hesamifard2017cryptodl} & 2, 3 & Full & Y & Y & N & N  \\ 
HCNN \cite{badawi2018towards} & 2  & Full & Y & Y & N & N  \\ 
DELPHI \cite{mishra2020delphi} & 2 & Partial & Y & Y & Y & N  \\ 
SAFENet \cite{SAFENET} & 2, 3  & Partial & N & Y & Y & N  \\ 
PreciseApprox \cite{lee2021precise} & 29  & Full & N & Y & N & Y$^*$ \\
QuaIL ({\bf Ours}) & 2 & Full & Y & Y & Y & Y  \\ \bottomrule
\end{tabular} 
}
\vspace{-2.2em}
\end{table}

The aforementioned challenges and inefficiencies of nonlinear computations using garbled circuits have driven researchers to design alternative activation functions that are cheaper to compute under PI. In particular, polynomial functions, which require only simple addition and multiplication, eliminate the need for garbled circuits and have become the de-facto solution for replacing ReLUs in neural networks. In fact, replacing all ReLUs with 
$x^2$ (denoted Quad here) 
can reduce online latency and communication dramatically by up to 2843$\times$ and 256$\times$, respectively \cite{mishra2020delphi}.

Table \ref{tab:MethodComph} summarizes prior work using polynomial activation functions for PI.
The partial/full distinction indicates whether the solution replaces some or all ReLU activations with polynomials.
We find that prior work can be classified into three categories:
full replacement using small datasets/models (e.g,. MNIST~\cite{mnist}, CIFAR-10~\cite{cifar})~\cite{gilad2016cryptonets,thaine2019efficient,chabanne2017privacy,mohassel2017secureml,fastercryptonets,hesamifard2017cryptodl,badawi2018towards}, 
partial replacement on mid-sized models (e.g., CIFAR-100)~\cite{mishra2020delphi,SAFENET}, 
and full-replacement on large models using very-high degree approximations~\cite{lee2021precise}.
Each of the solutions significantly advanced our understanding of the problem and the capabilities of PI.
However, none have demonstrated full replacement on large datasets/models using low-degree polynomials, which we believe is the ideal solution.




In this paper, we set out to replace all ReLUs with low-degree polynomials. Specifically, we test two drop-and-replace solutions (Taylor Approximation and Polynomial Regression Approximation) and develop two novel replace-and-retrain strategies (QuaIL and QuaIL+ApproxMinMax) on a wide range of networks and datasets. Our contribution can be summarized as follows:
\begin{enumerate}
\item We propose Quadratic Imitation Learning (QuaIL), a training setup inspired by dynamic programming to gradually build neural networks with only polynomial activations and introduce ApproxMinMaxNorm, a normalization strategy that bounds pre-activation values during training and approximately bounds pre-activation values during inference.
\item We implement and release Sisyphus, a set of methods for wholesale ReLU replacement that range from simple drop-and-replace solutions to replace-and-retrain strategies.
\item We develop and rigorously evaluate four substitution strategies using the Sisyphus framework and perform an in-depth analysis of their efficacy for deep networks. Crucially, we show that the instabilities of performing both inference and training with polynomial activation functions become more prominent in deeper neural networks and may not be observed in shallower networks.
\end{enumerate}
As we increase the complexity of the replacement strategy we steadily progress towards training deeper, more accurate, PI-friendly networks using only low-degree polynomial activations.
Despite our best efforts, we fall short of matching baseline ReLU network performance due to the \textit{escaping activation} problem: in all solutions, forward-pass activation values inevitably escape the well-behaving range of the polynomial activation function, leading to either exploding values (NaNs) or poor-behaving approximations.

Looking beyond QuaIL+ApproxMinMax (QuaIL+AMM), it may be tempting to evaluate additional solutions.
One way to overcome the escaping activation problem in QuaIL would be to bound the range.
However, this requires a max function, which if we had, we could simply use to compute ReLU in our networks.
Recent work proposed the Pade Activation Unit (PAU), 
a rational function of two low-degree polynomials that performs well on complex datasets~\cite{Molina2020Pade}.
Unfortunately, the division operation required by PAUs is not natively supported by cryptographic primitives of HE/SS and is known to be a challenge to implement.
Another recent work has proposed approximating ReLU and max-pooling using very-high (e.g., 29) degree polynomials and reports competitive accuracy for ImageNet \cite{lee2021precise}.
However, high-degree polynomials can be difficult to evaluate using cryptographic solutions as they would introduce significant additional computation in both SS and HE as well as noise growth in HE. 
Thus, we name this paper and our framework Sisyphus, as each time a promising solution was evaluated we incur a fundamental limitation that brought us back to square one.

\section{Methodology}
\label{sec:methodology}

We test the Sisyphus framework on the MNIST, CIFAR-10, CIFAR-100, and TinyImageNet \cite{yao2015tiny} datasets and test each substitution strategy over a wide variety of networks: AlexNet \cite{alexnet}, VGG11/VGG16 \cite{vgg}, ResNet18, MobileNetV1 \cite{howard2017mobilenets}, and ResNet32 \cite{he2016deep}. We develop and test our framework using PyTorch \cite{pytorch} (1.8.1+CUDA11.1), and for performing Bayesian Optimization during Polynomial Regression Approximation, we utilize GPyTorch \cite{gpytorch}, and BoTorch \cite{botorch}. 
All code for this paper is available 
online\footnote{See: \href{https://github.com/kvgarimella/sisyphus-ppml}{https://github.com/kvgarimella/sisyphus-ppml}}.

\section{Solutions and Results}
\label{sec:relwork}
In this section we present the solutions evaluated for replacing \textit{all} ReLUs with polynomial activation functions, 
including Taylor series approximation, polynomial regression, QuaIL, and QuaIL+AMM.

\begin{figure*}[ht!]
\begin{center}
\captionsetup{width=\textwidth}

\includegraphics[width=\textwidth]{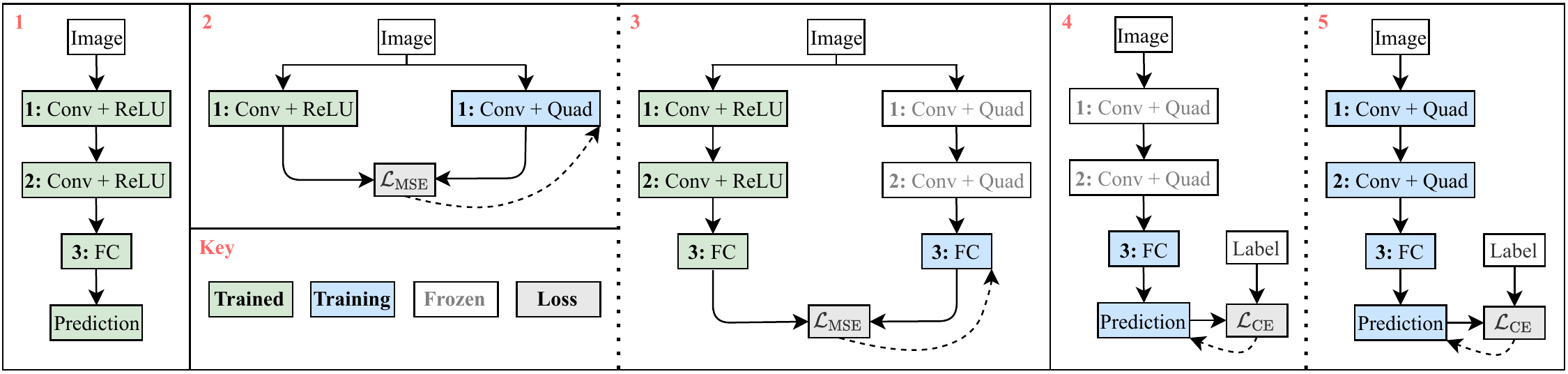} 

\centering{
  \caption{An overview of the \textbf{QuaIL} setup for a simple three-layer network. \textbf{1)} Train a baseline network with the ReLU activation function. \textbf{2)} Clone the first layer of the ReLU network, copy over the trained weights, and replace ReLU with Quad. Minimize the M.S.E. loss between the first-layer intermediate representations of the two networks and backpropagate through the Quad network. \textbf{3)} Repeat this process for each subsequent layer (while freezing the previous layers) until the full baseline network is cloned. \textbf{4-5)} Fine-tune the Quad network by gradually unfreezing layers and training with standard C.E. Loss. }\label{fig:quail_standard} }
\end{center}

\end{figure*}

\subsection{Drop-and-Replace}
\subsubsection{Taylor Series Approximation}
\label{sec:taylor}

\textbf{\\Key Idea}:
A simple approach to approximating ReLU as a polynomial is to use the Taylor Approximation. 
The Taylor approximation estimates a differentiable function, $f$, as a polynomial centered around point $a$ (we choose $a=0$). 
This approximation is constructed using high-order derivatives, and in the case of ReLU, all high-order derivative terms in the Taylor approximation vanish as the second derivative of ReLU is $0$ everywhere, resulting in a simple approximation: $\text{ReLU}(x) \approx \frac{1}{2}x.$

\noindent
\textbf{{Setup}}: 
First, a baseline ReLU model is trained.
We then replace all ReLUs in the trained networks with the Taylor approximation and measure the test accuracy for the network's respective dataset.

\noindent
\textbf{{Results}}:
As evident in Table \ref{tab:AccuracyResults}, the test accuracy deteriorates significantly for all networks except for the two layer MLP, which sees a dip in test accuracy from 97.98\% (using ReLU) to 86.28\% (using the Taylor approximation) on MNIST. Given that the Taylor approximation for ReLU is a simple linear function ($f(x) = \frac{1}{2}x$), we expect deeper networks to perform poorly when using the approximation as an activation function. 

\noindent
\textbf{{Takeaway}}: Using the Taylor approximation of ReLU collapses each network to a linear model, which restricts the network from representing the non-linear mappings required to achieve a high predictive performance on deeper networks and complex datasets.





\subsubsection{Polynomial Regression Approximation}
\label{sec:poly}

\noindent
\textbf{\\Key Idea}:
A natural extension of the Taylor approximation is to approximate ReLU using a polynomial over a range rather than a single point. 
The polynomial fit to a function $f$ has the form $\hat{f}(x) = \vec{w} \cdot \vec{x}$ , where $\vec{x} = (1,x^1,x^2,\dots,x^D )$ and
$D$ is the order of the polynomial. Polynomial regression can be employed to fit a polynomial function to any non-linear function by minimizing the mean squared error between the approximation and the target function $f$ over a range $[-a,a]$ and order $D$. 
For example if the target function is ReLU, optimal coefficients $w_0, w_1, \dots, w_D$ can be found by minimizing
\begin{equation}
\label{eq:poly_loss_func}
     E(\vec{w}) = \int_{-a}^{a}\left(\sum_{d=0}^{D}w_dx^d - \text{ReLU}(x) \right)^2 dx.
\end{equation}

\noindent
\textbf{Setup}: 
To find $\Vec{w}$ that minimizes Equation \ref{eq:poly_loss_func}, we first discretize the integral using a granularity of $dx=1e^{-3}$. 
The polynomial fit heavily depends upon the order of the polynomial ($D$) and the range ($a$) over which Equation \ref{eq:poly_loss_func} is minimized. To this end, we employ Bayesian Optimization (BayesOpt) to efficiently select effective values for $D$ and $a$ \cite{bayesopt}.
To accommodate a variety of polynomials, we choose the range $a$ to vary between $[0.5, 50]$ and the order of the polynomial to vary over integer values between $[2,9]$.


Given a setting of $\{D,a\}$, a ReLU approximation is found using polynomial regression.
All ReLUs in the original trained network are then replaced with the approximation and we measure the \textit{training} accuracy. 
BayesOpt uses this accuracy to iteratively update its probabilistic model and find well performing values of $a$ and $D$. 
We run BayesOpt for 50 iterations (10 random values to seed the probabilistic model and 40 optimized values) for each network and dataset. Finally, we replace all ReLUs in each network with the most accurate polynomial fit and measure the test accuracy.

\noindent
\textbf{Result}:
Table \ref{tab:AccuracyResults} displays the test accuracy for evaluating each network using the polynomial activation function produced by BayesOpt. Evaluating networks using non-linear polynomials introduces unbounded forward activations that compound exponentially with network depth, which we call \textit{escaping activations}. Especially for deeper networks, it is possible to generate forward activation values that overflow their floating point representations, which results in a NaN. We consider output logits that contain NaN values to be incorrect predictions. For this reason, two accuracies are presented in some rows of the polynomial regression experiments. Accuracies in parenthesis represent the test accuracy when only considering inputs that do not overflow in forward activation values. Using polynomial regression, we are able to progress to a high accuracy on LeNet, a five-layer network.

\noindent
\textbf{Takeaway}: Simply replacing all ReLUs with accurate polynomial approximations that are both low-degree and non-linear fails to work for most deeper networks due to the \textit{escaping activation} problem in which forward activation values grow exponentially, leading to instability in inference.

\subsection{Replace-and-Retrain}

\subsubsection{Quadratic Imitation Learning (QuaIL)}

\textbf{\\{Key Idea}}: The escaping activation problem encountered when using the polynomial regression strategy was directly caused by the compounding use of polynomials in deeper networks. Specifically, after each pass through the polynomial activation, the output \textit{intermediate representation values} began to grow exponentially. Following several layers (several passes through the polynomial activations), the intermediate representation values escaped the well-behaving regions of the polynomials and resulted in exploding values (NaNs). The escaping nature of intermediate representations suggests to elevate from simple drop-and-replace strategies to replace-and-retrain strategies which mitigate the escaping activation problem. Rather than attempting to train a network with polynomial activations end-to-end by minimizing the loss between ground truth and predictions, Quadratic Imitation Learning (QuaIL) iteratively builds and trains a neural network with polynomial activations by mimicking the intermediate representation values of a trained ReLU network. Similar to dynamic programming, QuaIL attempts to first solve a sub-problem by mimicking intermediate representation values of a well-behaving network before adding additional layers to a  network using polynomial activations. In QuaIL, the polynomial activation function is set to $f(x) = x^2$ (Quad). 
\textbf{\\{Setup}}:
Figure \ref{fig:quail_standard} depicts the QuaIL training process. First, a ReLU baseline network is trained using standard supervised learning techniques (Fig. ~\ref{fig:quail_standard}.1). 
Then, the first layer of the ReLU network is duplicated and the layer's ReLUs are replaced with Quad. 
Here, the Quad network is trained by minimizing the Mean Square Error (M.S.E.) between the first-layer intermediate representations of both networks (Fig. ~\ref{fig:quail_standard}.2). In this way, the single-layer Quad network learns to predict similar first-layer representations as the ReLU network. After training converges to a low M.S.E. between the two intermediate representations, the Quad network's first-layer weights are frozen and the second layer of the ReLU network is cloned and stacked onto the Quad network. Again, ReLU is replaced by Quad for the second layer. Similar to the first layer, the Quad network now minimizes the M.S.E. between the second-layer representations of both networks. This process is repeated until the final layer of the ReLU network has been added to the Quad network and the error between the final representations is minimized (Fig. ~\ref{fig:quail_standard}.3). 

At this stage, the Quad network is trained using standard supervising learning while gradually unfreezing shallower layers. In the image classification setting, the Cross Entropy (C.E.) loss is minimized between ground truth labels and predictions (Fig. ~\ref{fig:quail_standard}.4-~\ref{fig:quail_standard}.5).

\begin{table}[!htbp] 
\caption{Accuracy when all ReLUs in the networks are substituted with Taylor series approximation (T-Approx.), polynomial regression approximation (Poly-Reg.), QuaIL, and QuaIL+AMM on CIFAR-10/100 (C-10/100) and TinyImageNet (Tiny) datasets.}
\label{tab:AccuracyResults}
\resizebox{0.48\textwidth}{!}{
\begin{tabular}{p{2.5cm}|r|r|r|r|r} \toprule
Dataset-Net & \multicolumn{1}{l|}{\begin{tabular}[c]{@{}l@{}}Baseline \end{tabular}} & \multicolumn{1}{l|}{\begin{tabular}[c]{@{}l@{}}T-Approx. \end{tabular}} & \multicolumn{1}{l|}{\begin{tabular}[c]{@{}l@{}}Poly-Reg.\end{tabular}} & \multicolumn{1}{l|}{\begin{tabular}[c]{@{}l@{}}Quail \end{tabular}} & \multicolumn{1}{|l}{\begin{tabular}[c]{@{}l@{}} Quail+AMM \end{tabular}} \\ \toprule
MNIST-MLP & 97.98 & 86.28 & 97.88 & 98.20 & 98.17 \\
MNIST-LeNet & 99.32 & 9.81 & 99.14 & 99.45 & 99.26 \\ \midrule
C10-AlexNet & 85.47 & 12.90 & 1.28 (71.11)  & 79.94 & 79.32 \\
C10-VGG11 & 90.46 & 13.68 & 8.49 (82.83) & 82.19 & 82.85 \\
C10-VGG16 & 92.78 & 16.01 &  13.31 (87.57)& --- & 82.25 (82.63) \\
C10-ResNet18 & 93.21 & 1.00 & 13.64 & --- &  83.61 (85.72)\\
C10-MobileNetV1 & 91.70 & 11.04 & 11.13 & 10.00 & 48.03 (49.45) \\
C10-ResNet32 & 91.72 & 26.96 & 90.48 (90.62) & --- & 56.93 (71.81)\\ \midrule
C100-AlexNet & 60.98 & 1.73 & 21.51 (31.8)& 54.83 & 50.76 \\
C100-VGG11 & 68.36 & 1.88 & 3.74  & 52.08 & 55.53 \\
C100-VGG16 & 71.44 & 1.74 & 0.94 (52.81) & --- & 54.56 (55.03) \\
C100-ResNet18 & 74.39 & 1.00 & 1.00 & --- & 65.17 (66.30)\\
C100-MobileNetV1 & 65.00 & 1.00 & 25.8 & --- & 0.92 (01.09)  \\
C100-ResNet32 & 67.83 & 4.28 & 66.31 & --- & 19.86 (28.58) \\ \midrule
Tiny-AlexNet & 51.65 & 0.51 & 1.70 (8.98) & 38.95 & 36.24 \\
Tiny-VGG11 & 55.36 & 0.66 & 0.5 & 37.59 & 44.63 (44.68) \\
Tiny-VGG16 & 58.88 & 0.45 & 00.56 (3.07)  & --- &  45.76 (46.47)\\
Tiny-ResNet18 & 61.59 & 0.50 & 0.5 & --- &  49.45 (53.82) \\
Tiny-MobileNetV1 & 56.16 & 0.76 & 0.46 & --- & 13.43 (22.75)  \\
Tiny-ResNet32 & 54.77 & 2.04 & 47.55 & --- & 7.16 (10.60)  \\ \bottomrule
\end{tabular}}
\end{table}

\noindent
\textbf{Result}:
QuaIL further extends our progress of building deep, PI-friendly networks to AlexNet and VGG11 on the CIFAR-10, CIFAR-100, and TinyImageNet datasets. Each of these networks that \textit{only uses the} Quad activation function are built up iteratively to limit the effect of \textit{escaping activations}. However, QuaIL fails to generalize to even deeper networks as even a small difference in the intermediate representations at earlier stages of the deeper networks propagate forward leading to \textit{escaping activations} and causing training to diverge (denoted as --- in Table \ref{tab:AccuracyResults}).

\noindent
\textbf{Takeaway}: QuaIL allows us to iteratively build deep (up to 11-layer) networks with only the Quad activation function but fails to mitigate unstable intermediate representations for even deeper networks and thus still suffers from \textit{escaping activations}. For example, a ResNet18 network trained using the QuaIL setup experiences exploding gradients due to escaping activations in latter intermediate representations and is unable to converge during training.

\subsubsection{Approximate MinMax Normalization}
\textbf{\\{Key Idea}}:
The escaping activation problem encountered during QuaIL illustrates the need to bound pre-activation values to train networks using low-degree polynomial activation functions, especially for deeper neural networks. 
To do this we developed Approximate Min-Max Normalization (ApproxMinMaxNorm), which places upper and lower constraints on pre-activation values during training by performing a dimension-wise Min-Max normalization:
\begin{equation}
    \hat{x} = \alpha \frac{x - \min x}{\max x - \min x} - \beta
\end{equation}
where $\alpha$ and $\beta$ are scaling parameters. During the training phase, approximations of minimums and maximums are calculated and stored using a weighted moving average of the true minimums and maximums (we use a smoothing factor of $1/10$). When performing inference, these stored approximations are then used to perform approximate  normalization.
\textbf{\\{Setup}}:
ApproxMinMaxNorm is combined with the QuaIL training procedure; when building the Quad network, ReLU is replaced by an ApproxMinMaxNorm layer immediately followed by Quad. 
\textbf{\\{Result}}:
We observe stable training for \textit{all} networks and datasets using QuaIL+AMM. However, at inference time, when using the approximated values of the minimums and maximums for each layer, we again detect the escaping activation problem, albeit to a less degree when compared to the drop-and-replace polynomial regression strategy. 
\textbf{\\{Takeaway}}:
ApproxMinMaxNorm prevents the \textit{escaping activation} problem at training time by explicitly bounding the pre-activation values to polynomial activations. However, the escaping activation problem returns during inference due to approximate minimum and maximum calculations. Thus \textit{a true maximum function} is required at test time to guarantee bounds on pre-activation values.



\section{Discussion}
\label{sec:discussion}
The desirable properties of PI-friendly ReLU substitutions are: low multiplicative depth, stability over a sufficiently large range of activation values, and  competitive performance when compared to networks with ReLUs. The Quad activation function has been considered a promising solution as its multiplicative depth is one and exhibits stability for simple models and datasets, e.g., MNIST \cite{gilad2016cryptonets}. However, for deeper networks and larger datasets, the desired stability range of pre-activation values increases significantly \cite{lee2021precise} and using Quad in this extended range results in imprecise approximations of ReLU and poor accuracy \cite{chabanne2017privacy}. Consequently, higher-degree polynomials are used for more accurate approximation, but suffer from a higher multiplicative depth that results in additional computation (in SS/HE) and noise growth (HE), thus limiting their efficacy in practical settings. 


To help mitigate these issues, we devised QuaIL where each layer in the Quad-network learns to mimic intermediate representations of a trained, all-ReLU network. QuaIL worked well for AlexNet and VGG11, which polynomial regression under-performed; however, it did not scale to even deeper networks. To understand why, we dug deeper and found the issue still to be \textit{escaping activations}. That is, some intermediate representation values still began to grow unbounded.
\begin{figure}[t]
\begin{center}
\includegraphics[scale=0.45]{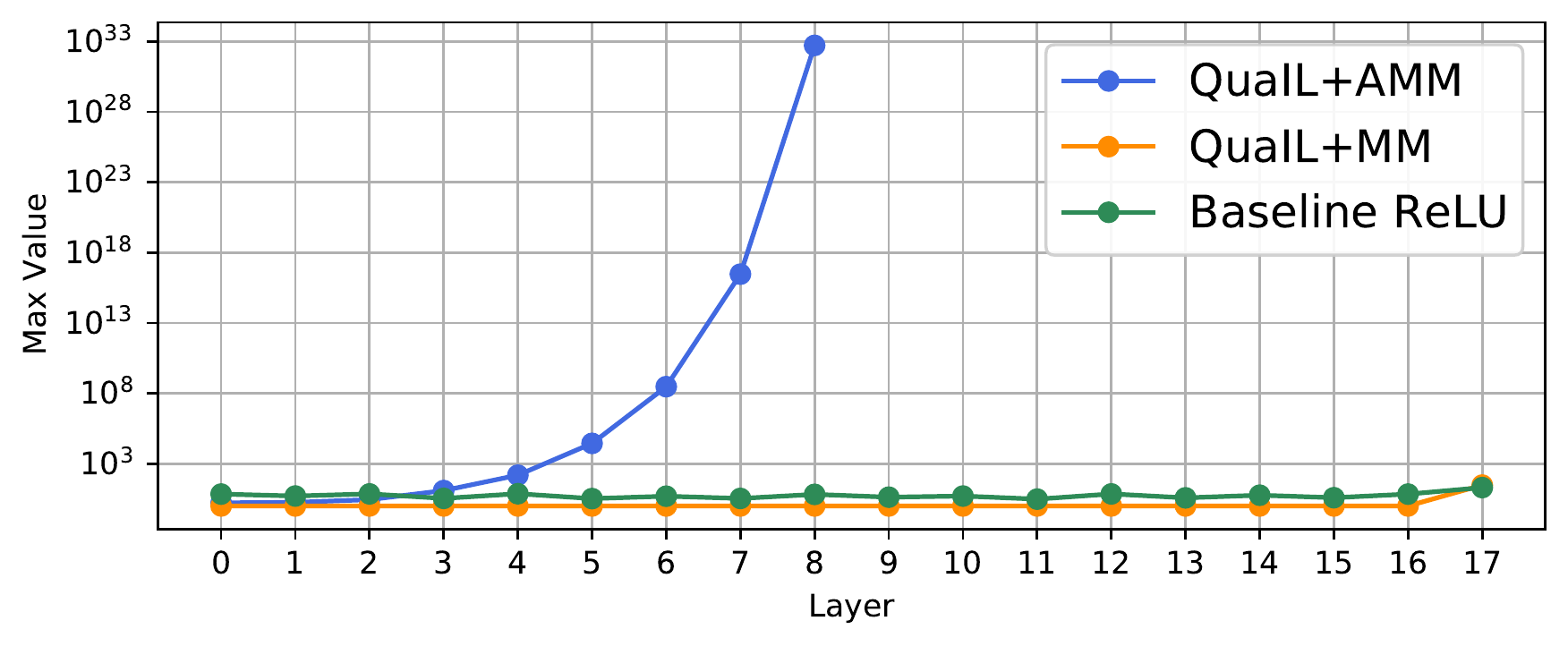}
\end{center}
\vspace{-1em}
\caption{Comparison of maximum forward activation values after each nonlinear layer at inference time using Quail+AMM (ApproxMinMax), Quail+MM (TrueMinMax), and an all-ReLU baseline ResNet18.}
\label{fig:escaping}
\end{figure}

 To mitigate \textit{escaping activations} at training time, we bounded the pre-activation values inputs using an ApproxMinMax normalization strategy, which achieved reasonable accuracies for all the networks except MobileNetV1 and ResNet32. However since the maximum and minimum values were approximated at inference time, the approximation error grew in deeper layers and some activations began to explode (shown in Figure \ref{fig:escaping} as Quail+AMM). For a better understanding, we replaced the approximated min and max with the true min and max during inference (termed as QuaIL+MM in Figure \ref{fig:escaping}) and observed that the intermediate representation values were now similar to that of the all-ReLU baseline networks. 

Fundamentally, the efficacy of using low-degree polynomials for deeper networks on complex datasets boils down to bounding input values to the the polynomial activations in order to mitigate \textit{escaping activations}, which requires using exact calculations of both the minimum and maximum. However, the issue of calculating exact minimums and maximums brings us back full circle to the problem we were trying to solve: remove all ReLUs (which is defined using maximum) to prevent the usage of garbled circuits in PI. We hope the insights gained from Sisyphus aid the PPML community in being mindful when using low-degree polynomial activations in PI-friendly networks.


\section*{Acknowledgements}
\label{sec:ack}
This work was supported in part by the Applications Driving Architectures (ADA) Research Center, a JUMP Center co-sponsored by SRC and DARPA. This research was
also developed with funding from the Defense Advanced
Research Projects Agency (DARPA),under the Data Protection in Virtual Environments (DPRIVE) program, contract
HR0011-21-9-0003. The views, opinions and/or findings
expressed are those of the author and should not be interpreted as representing the official views or policies of the
Department of Defense or the U.S. Government.

\bibliographystyle{ACM-Reference-Format}
\bibliography{References}


\end{document}